\documentclass[sigconf]{acmart}

\usepackage{multirow}
\usepackage{amsmath}
\usepackage{booktabs}
\usepackage{array}

\AtBeginDocument{%
  }

\setcopyright{acmlicensed}
\copyrightyear{2024}
\acmYear{2024}
\acmDOI{10.1145/3669828.3669835}

\acmConference[IMIP2024]{International Conference on Intelligent Medicine and Image Processing}{April 26--29,
  2024}{Bali, Indonesia}
\acmISBN{979-8-4007-1003-2/24/04}

\begin{document}

\title{Toward an Integrated Decision Making Framework for Optimized Stroke Diagnosis with DSA and Treatment under Uncertainty}

\author{Nur Ahmad Khatim}
\affiliation{%
  \institution{Institut Teknologi Sepuluh Nopember (ITS)}
  \city{Surabaya}
  \country{Indonesia}
}
\author{Ahmad Azmul Asmar Irfan}
\affiliation{%
  \institution{Universitas Islam Negeri (UIN) Syarif Hidayatullah}
  \city{Jakarta}
  \country{Indonesia}
}

\author{Amaliya Mata'ul Hayah}
\affiliation{%
  \institution{Universitas Islam Negeri (UIN) Syarif Hidayatullah}
  \city{Jakarta}
  \country{Indonesia}
}

\author{Mansur M. Arief}
\affiliation{%
  \institution{Stanford University}
  \city{Stanford}
  \state{California}
  \country{USA}
}
\email{mansur.arief@stanford.edu}







\renewcommand{\shortauthors}{Khatim et al.}

\begin{abstract}
  This study addresses the challenge of stroke diagnosis and treatment under uncertainty, a critical issue given the rapid progression and severe consequences of stroke conditions such as aneurysms, arteriovenous malformations (AVM), and occlusions. Current diagnostic methods, including Digital Subtraction Angiography (DSA), face limitations due to high costs and its invasive nature. To overcome these challenges, we propose a novel approach using a Partially Observable Markov Decision Process (POMDP) framework. Our model integrates advanced diagnostic tools and treatment approaches with a decision-making algorithm that accounts for the inherent uncertainties in stroke diagnosis. Our approach combines noisy observations from CT scans, Siriraj scores, and DSA reports to inform the subsequent treatment options. We utilize the online solver DESPOT, which employs tree-search methods and particle filters, to simulate potential future scenarios and guide our strategies. The results indicate that our POMDP framework balances diagnostic and treatment objectives, striking a tradeoff between the need for precise stroke identification via invasive procedures like DSA and the constraints of limited healthcare resources that necessitate more cost-effective strategies, such as in-hospital or at-home observation, by relying only relying on simulation rollouts and not imposing any prior knowledge. Our study offers a significant contribution by presenting a systematic framework that optimally integrates diagnostic and treatment processes for stroke and accounting for various uncertainties, thereby improving care and outcomes in stroke management.
\end{abstract}





\maketitle

\section{Introduction}
Accurate diagnosis and treatment of stroke are crucial due to the condition's rapid progression and potentially severe consequences. Stroke, often resulting from conditions such as aneurysms, arteriovenous malformations (AVM), or occlusions, necessitates swift and precise medical intervention. Approximately 87\% of all strokes are ischemic, with an age-standardized incidence of cerebral infarction at 153.2 per 100,000 person-years. Additionally, the age-standardized incidence of hemorrhagic stroke is 46.2 per 100,000 person-years. Diseases associated with hemorrhagic stroke, like intracerebral aneurysm, have an incidence rate of 12.6 per 100,000 person-years, while AVM incidence stands at 2.4 per 100,000 person-years \cite{lee2020trends, duloquin2020incidence}.

The current approach to these conditions includes diagnostic imaging, such as Digital Subtraction Angiography (DSA), coupled with individualized treatment plans. DSA is a key imaging modality for identifying and confirming aberrations in cerebral blood vessels in various conditions, including aneurysms and arteriovenous malformations. It is considered the gold standard for evaluating cerebral vascular disease and is crucial for treatment planning in patients with suspected vascular abnormalities \cite{settecase2021advanced}.

In addition, several stroke assessment tools have been developed for rapid diagnosis and triage. These tools facilitate the quick identification of acute ischemic stroke and are validated for clinical use. However, they come with significant costs, and false positives can place additional strain on the healthcare system. An evaluation of 13 clinical measures for identifying occlusion in over 1,000 individuals revealed high rates of false negatives and positives \cite{turc2016clinical}. Furthermore, a meta-analysis indicated that no scale could accurately and consistently identify occlusion \cite{patil2022detection}, underscoring the challenges in stroke diagnosis. This is mostly because many aspects of stroke may not be directly observable, such as microvascular changes or silent strokes, which lack apparent symptoms \cite{vermeer2007silent}.

In this complex and highly uncertainty scenario, healthcare professionals often must make critical decisions based on partial information, compounded by the inherent history and physical examination noise in diagnostic processes. This often leads to uncertainties in diagnosis and treatment, where every decision carries significant weight. The consequences of incorrect decisions extend beyond immediate healthcare costs; they profoundly impact patient quality of life, and long-term health outcomes \cite{donkor2018stroke}. Under such a setting, an intelligent decision-making framework that we propose offers a transformative solution. By integrating various diagnostic tools and standardized treatment protocol with algorithms for optimizing decisions, we can incorporate the risks associated with partial information and observation noise into an uncertainty-aware decision. This approach echoes earlier works that have leveraged systematic algorithms and optimization models to address similar challenges and incorporates decision making under uncertainty, such as Markov Decision Processes (MDPs) or Partially Observable MDPs (POMDPs) \cite{steimle2017markov, tsoukalas2015data, schell2016data}.

MDPs and POMDPs are particularly adept at modeling decision-making scenarios where outcomes are partly random and partly under the control of a decision-maker \cite{kochenderfer2022algorithms}. MDPs, for instance, have been effectively used in scenarios requiring full state observability, assuming a 100\% accurate diagnosis a priori. The strength of these models, especially in offline algorithms, lies in their ability to solve to optimality given a sufficient compute time. However, they are often constrained by the curse of dimensionality (i.e. being exponentially slower as the problem size increases) and full observability requirement. POMDPs extend the capabilities of MDPs by accounting for partial observability, making them more suitable for medical scenarios where diagnosis and patient responses to treatment can be uncertain and variable. This is especially important when it comes to stroke care, where quick and precise decision-making is critical. POMDP assists in reaching a compromise between the time and resource limitations and the requirement for comprehensive diagnostics \cite{ehrmann2023making}. 

\begin{figure*}
    \centering
    \includegraphics[width=0.7\linewidth]{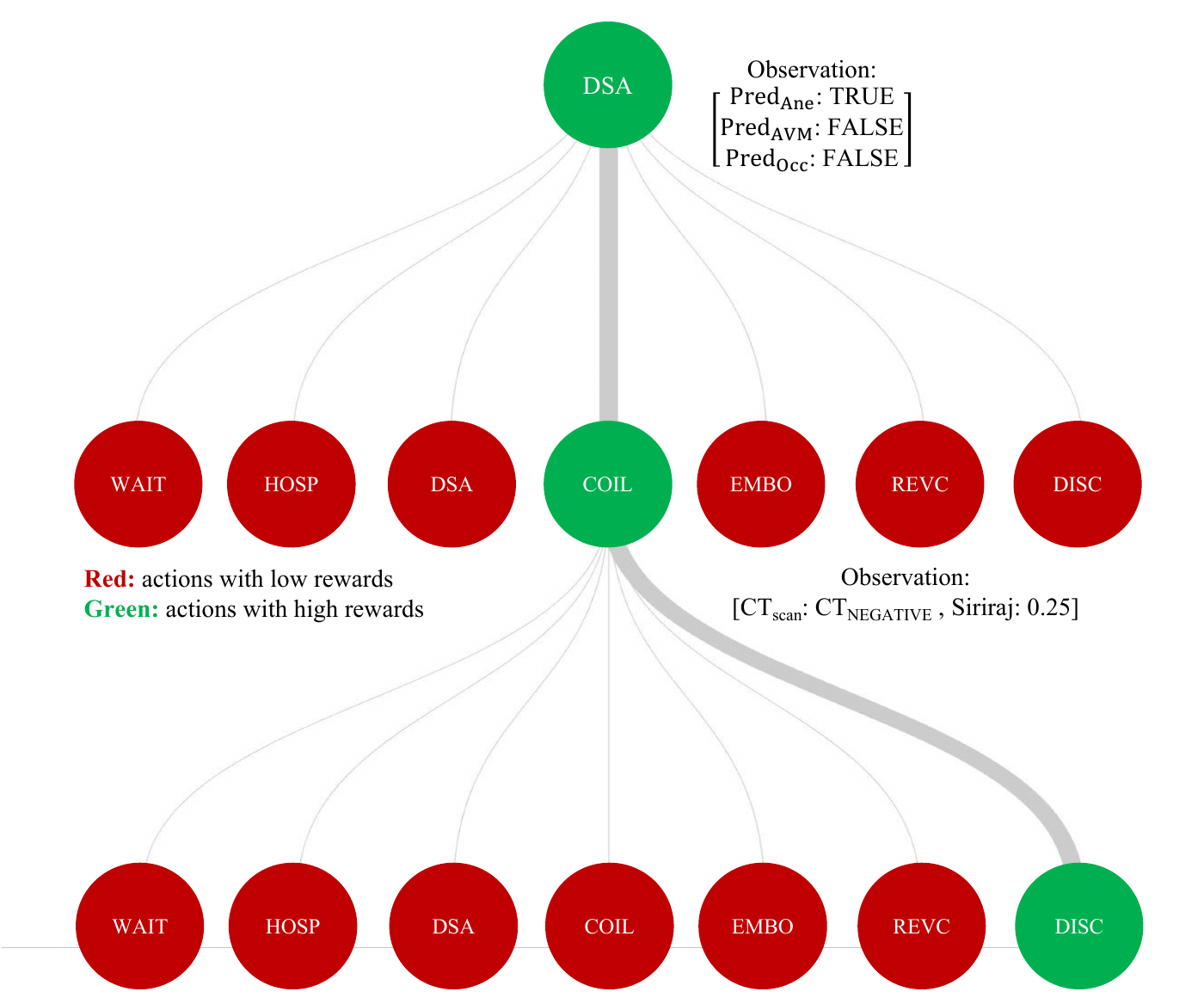}
    \caption{Illustration of simulated policy rollouts in tree search}
    \label{fig:rollouts_illustration}
\end{figure*}

POMDP is often solved using online algorithms. A key feature of online algorithms is their use of tree-search methods combined with particle filters \cite{fischer2020information}, which enable the simulation of future trajectories. This approach simulates future observations and trajectories as a tree when a certain action is taken and utilizes these simulated rollouts to make informed decisions on the possible future progression of the states (i.e. aneurysm, AVM, or occlusion cases in our context). This procedure is often illustrated in tree diagrams (see Fig. \ref{fig:rollouts_illustration}).

Despite their potential to significantly enhance the accuracy and efficiency of medical interventions in complex cases, there is a notable lack of studies exploring the use of online algorithms in the diagnosis and treatment of aneurysm, AVM, or occlusion stroke conditions. In this study, we propose a POMDP formulation that allows us to consider both diagnosis and treatment simultaneously. More importantly, the important feature that we aim to capture is the limited information during the diagnosis stage and the impact of imperfect diagnosis on the optimality of the treatment. To study this, we build a model that does not assume full information of the stroke type, but only noisy observations in the form of CT scan reading, Siriraj score, or DSA report, with varying sensitivity in identifying the occurrence of aneurysm (96.7\%), AVM (100\%), or occlusions (96\%) \cite{feng2020diagnostic, chowdhury2015digital, liu2020comparison}. We then compare the performance of various baselines including random policy, expert policy, and optimized policies using an online solver DESPOT \cite{ye2017despot}. 

In summary, our contribution is twofold. First, we propose a POMDP formulation for integrated stroke diagnosis and treatment for long-horizon problems. Second, we provide a benchmark of our POMDP policy with expert policies under settings with high uncertainty to mimic real-world scenarios and stress-test the usefulness of these algorithms. The code for our experiments is available at: \href{www.github.com/inteligensi/DSAPOMDPs.jl}{www.github.com/inteligensi/DSAPOMDPs.jl}.

The rest of this paper is organized as follows. In Section \ref{sec:literature_review}, we review works that address stroke diagnosis and introduce POMDP framework. In Section \ref{sec:problem_formulation}, we present our model and approach to solving it using online solver. We then provide details of our experiments in Section \ref{sec:exp}. Finally, we discuss our findings in Section \ref{sec:discussion} and conclude in Section \ref{sec:conclusion}.

\section{Related Work}
\label{sec:literature_review}

In this section, we briefly summarize stroke diagnosis and treatment, POMDP, and its uses for assisting medical decision making. 

\subsection{Stroke Diagnosis and Treatment}

Swift decision-making is crucial in stroke management, with significant implications for patient outcomes. Rapidly identifying symptoms and determining their cause is essential for effective treatment and minimizing long-term effects. Medical imaging, especially CT scans and DSA, is vital in this process. CT scans are key for early stroke diagnosis, helping to identify ischemic or hemorrhagic areas by visualizing changes in brain tissue density. DSA, on the other hand, is the gold standard for evaluating and treating neurovascular diseases like AVM and aneurysms in hemorrhagic strokes \cite{settecase2021advanced}, though highly invasive and costly in nature and thus should only be ordered only when necessary.

These diagnostic procedures is highly important to guide effective treatments. For instance, approaching ischemic stroke cases with quick revascularization, such as thrombolysis, has been associated with better outcomes. AVM treatments range from endovascular embolization, a minimally invasive procedure, to microsurgical resection \cite{derdeyn2017management}. Intracranial aneurysms may require transluminal coiling, a procedure using a catheter to prevent blood flow into the aneurysm \cite{zhao2018current}. These treatments, however, often have a limited window of opportunity, and delays can impact patient eligibility and prognosis \cite{hurford2020diagnosis}. Thus, integrating appropriate diagnostics and treatments is essential for resource optimization, expertise application, and patient-specific care, addressing the inherent uncertainties of stroke management.

Addressing these uncertainties is challenging. Stroke symptom variability, patient comorbidities, and individual factors often lead to incomplete and ambiguous information. Diagnostic noise, such as imaging artifacts, complicates symptom interpretation \cite{merino2013predictors}, making accurate and timely diagnosis difficult. Healthcare professionals must often make critical decisions quickly, considering available resources. To mitigate this, patients are frequently observed in hospitals to accumulate lab measurements and physical observations. Repeated close observation of physical and physiological symptoms can build diagnostic confidence, albeit time-consuming. The Siriraj score \cite{poungvarin1991siriraj} is commonly used to summarize these observations, guiding treatment or follow-up procedures. Scores below -1 typically indicate ischemic stroke, while scores above +1 suggest hemorrhagic stroke. Inconclusive results, around 0, often lead to more advanced diagnostics like DSA. Combining these procedures, healthcare professionals must navigate uncertainties to make critical decisions, often life-altering, and provide the best possible medical advice for patient treatment.

\subsection{Partially Observable Markov Decision Process (POMDP)}

POMDPs provide a robust mathematical framework for modeling decision-making problems under uncertainty and partial observability. In a POMDP, an agent interacts with an environment where the true state is not directly observable. The goal is to make decisions that maximize the expected cumulative reward over time \cite{smallwood1973optimal}. The key challenge is to maintain a belief state—a probability distribution over possible states—based on observed information and to optimize decisions amidst uncertainty \cite{madani1999undecidability}.

Given the inherent complexity of POMDPs, online solvers have become essential for real-time decision-making. These solvers adapt dynamically to changing environments by continuously updating decision strategies based on observed information \cite{ross2008online}. Online planning involves simultaneous planning and execution. At each time step, it conducts local planning, selecting the best action based on the current belief and employing a lookahead search within the vicinity of the existing belief state. The chosen action is then executed promptly. This approach offers computational benefits: it is less complex to search for an optimal action within a single belief than for all beliefs, as required in offline policy computation. Additionally, it leverages local structures, reducing the search space size. However, the online approach is limited by the available planning time.

A notable online solver for POMDPs, particularly in decentralized multi-agent settings, is the Determinized Sparse Partially Observable Tree (DESPOT) \cite{ye2017despot}. This algorithm adapts the Monte Carlo Tree Search \cite{winands2008monte} framework to decentralized scenarios, where multiple agents with partial observability must coordinate actions to achieve a common goal. DESPOT maintains belief states for each agent using particle filters and employs an adaptive tree depth strategy to optimize computational resources. By using rollout policies and approximating the value function through MCTS, DESPOT excels in real-time decision-making under uncertainty. DESPOT's decentralized structure and adaptability make it effective for large-scale problems where computational efficiency is crucial. The solver has been applied in research scenarios, such as addressing POMDP challenges in Highway Driving \cite{ulfsjoo2022integrating}.

\subsection{POMDP for Medical Decision Making}

Decision-making challenges in stroke management often arise from various factors, including subtle symptoms and the variability of treatment options within a time-sensitive framework. Implementing POMDP in the context of stroke events offers potential benefits, particularly in optimizing diagnostic and treatment decisions. While specific studies on this topic may be limited, the application of POMDP in healthcare has shown promise.

In the context of diagnosis, an earlier study explored the timing of mammograms for suspected breast cancer cases. The researchers developed a finite-horizon POMDP model to determine personalized mammography screening policies based on a woman’s risk factors and past screening results. The model's unobservable states represented the patient’s cancer stage, ranging from noninvasive to invasive, under-treatment, or no cancer. The POMDP actions were “wait” and “mammography,” with outcomes including positive or negative mammogram results, or self-detection results. Positive findings led to further diagnostic steps like biopsies. The study revealed a control-limit policy based on the risk of invasive and noninvasive cancers, influenced by the patient's screening history~\cite{ayer2012or}.

In treatment contexts, another study developed a framework for personalizing anticoagulant therapy in patients on warfarin. The POMDP model was used to update beliefs and assess the optimality of myopic policies in MDP, deriving conditions for their existence and uniqueness. Real-life patient data validated the model, offering practical clinical insights, such as medication initiation duration and patient medication sensitivity \cite{ibrahim2016designing}.

Our study proposes a POMDP formulation for integrated stroke diagnosis and treatment. The goal is to combine diagnosis and treatment decision-making and assess how the model, solved with an online solver, performs under various initial that mimic real world. This approach will be compared with heuristic policies currently in standard practice, such as hospital observation or direct DSA ordering. We hypothesize that our POMDP approach can provide a decision support system, offering healthcare professionals recommendations to navigate partial observation, state uncertainty, and the significant consequences of decision-making in these settings.

\section{Methodology}
\label{sec:problem_formulation}

In this section, we present the formulation of our problem as a POMDP. More generally, a POMDP model is defined by the tuple $(\mathcal S, \mathcal A, \mathcal O, T, Z, R, \gamma)$, where $\mathcal S$ is the state space, $\mathcal A$ is the action space, and $\mathcal O$ is the observation space. For $s, s' \in \mathcal S, ~a \in \mathcal A, ~T(s'~|~s, a)$ is the probability of transitioning into state $s'$ from current state $s$ by taking action $a$. $Z(o~|~s, a)$ is the probability of observing observation $o \in \mathcal O$ given that at state $s$ action $a$ is taken. $R(s, a)$ is the immediate reward obtained when taking action $a$ in state $s$. Finally, $\gamma \in (0, 1)$ is the discount factor to incorporate how much we care about future rewards.

Specifically, for our problem, the state 
\begin{align}
s \in (isAne, isAVM, isOcc, t)
\end{align}
contains the patient’s stroke-related condition, including binary variables that indicate the occurrence of aneurysm ($\text{is}_{{Ane}}$), AVM ($isAVM$), and occlusion ($isOcc$). We also store as part of the state a non-negative discrete time counter $t$. The action space $A$ includes both diagnostic, i.e. observe closely in a hospital (HOSP) or perform DSA (DSA) and surgical treatment options that a medical team can take for the patient, including coiling (COIL), embolization (EMBO), or revascularization (REVC) \begin{align}
A = \left( \text{WAIT, HOSP, DSA, COIL, EMBO}, \text{REVC, DISC} \right). \nonumber
\end{align} 

In addition, we also add a discharge action (DISC) to signify when the patient has been treated and can be sent home. Note here that the action WAIT signifies the case in which the patient's condition is deemed healthy enough for the time being to wait at home without treatment just yet.

The defining characteristic of POMDP is the partial unobservability of the states. For our problem, we do not have full access to $s_{t} \in \mathcal S$, the state at time $t$, i.e. we do not know $isAne$, $isAVM$, or $isOcc$, but only know $t$. However, we can observe $o_{t} \sim Z( . ~| ~s_{t}, a_{t}) \in \mathcal O$ that provides some (noisy) lab measurements or observations about the patient that can be used to build a belief model $b_{t}$ to estimate the underlying value of $s_{t}$. It is based on this belief $b_{t}$ that we can then take actions $a_{t} = \pi(b_{t}) \in \mathcal A$ based on some policy $\pi$, after which a reward $r_{t} = R(s_{t}, a_{t})$ is awarded based on the current state $s_{t}$ and selected action $a_{t}$ and the system progresses into the next state   $s_{t+1} \sim T( . ~| ~s_{t}, a_{t})$. It is worth noting here that the original state $s_{0} \sim p_{0}$ is often treated as a random variable sampled from some initial state distribution $p_{0}$. For our problem, $p_{0}$ is an estimate of how the general population encounters stroke conditions and seeks medical help.

To solve the formulated POMDP, a solver seeks an optimal policy $\pi^{*}$ that maximizes the sum of discounted rewards (often referred to as utility) \begin{align}
\pi^{*} = \arg \max_{\pi } ~\mathbb E \left[ \sum_{t=1}^{\infty} \gamma^t r^t \right].
\label{optimal_policy}
\end{align} It should be clear from (\ref{optimal_policy}) that the complexity of the problem is heavily influenced by the combination of possible values that $s_{t}$, $a_{t}$, and $o_{t}$ may take (the size of $\mathcal S$, $\mathcal A$, and $\mathcal O$), as well as the complexity of the reward function $R$ and policy space. 

To better describe our modeling approach, below is a more detailed description of $T$ (transition model), $Z$ (observation model), and $b_{t}$ (belief model).

\subsection{Transition Model}

The transition model T represents how the patient’s state changes over time and incorporates both the effect of the treatment-type actions (COIL, EMBO, REVC) on the state as well as naturalistic randomness of the world (random or unmodeled occurrence of stroke).

For simplicity, we assume each of the treatment-type actions treats one specific type of stroke condition (COIL treats aneurysm, EMBO treats AVM, and REVC treats occlusion) with probability 1, and it does not have any effect on other stroke conditions. We can extend our model to allow treatments affecting multiple conditions. However, it would require more parameters to estimate a priori without adding much insights into the usefulness of our study significantly. We thus omit such an effect in the current study.

In addition to treatment effects, we also model the plausibility of random (unmodeled) stroke occurrence in healthy patients. To account for this, we model such occurrence events as independent Bernoulli trials at each time step with probability $p_{ane}$, $p_{AVM}$, and $p_{occ}$ for aneurysm, AVM, and occlusion, respectively. Finally, the time counter t  is updated at every time step, regardless of actions. Table \ref{tab:transition_probability} summarizes the state-transitioning events with non-zero probability.

\begin{table}
  \caption{Non-Zero State Transition Probability}
  \label{tab:transition_probability}
  \centering
  \begin{tabular}{llc}
    \toprule
    Action & Transition Events & Probability \\
    \midrule
    COIL & $IsAne: TRUE \rightarrow IsAne: FALSE$ & 1.0 \\
    EMBO & $IsAVM: TRUE \rightarrow IsAVM: FALSE$ & 1.0 \\
    REVC & $IsOcc: TRUE \rightarrow IsOcc: FALSE$ & 1.0 \\
    \midrule
    \multirow{4}{*}{Any $a \in \mathcal{A}$} & $IsAne: FALSE \rightarrow IsAne: TRUE$ & $p_{ane}$ \\
                                             & $IsAVM: FALSE \rightarrow IsAVM: TRUE$ & $p_{AVM}$ \\
                                             & $IsOcc: FALSE \rightarrow IsOcc: TRUE$ & $p_{occ}$ \\
                                             & $Time: t \rightarrow Time: t + 1$ & 1.0 \\
    \bottomrule
  \end{tabular}
\end{table}

\subsection{Observation Model}

The observation model $Z$ encapsulates the processes of lab measurement, patient observation, and report summarization as part of the treatment process. It assumes that certain symptoms will manifest in lab measurements (such as CT scan readings), patient conditions (summarized as the Siriraj score), or angiography procedures (like DSA), signaling the underlying stroke conditions ($IsAne$, $IsAVM$, $IsOcc$), albeit with some level of noise.

For instance, a CT scan provides imaging results that the medical team interprets to determine the presence of any of the three stroke conditions. However, these conditions may not always be distinguishable solely from the scans. If a patient's condition is ($isAne: TRUE, isAVM: FALSE, isOcc: FALSE$), the CT scan summary might indicate $CT\_POSITIVE$ with a probability p1 and $CT\_NEGATIVE$ with a probability $1 - p_{1}$. These probabilities vary based on the patient's specific condition. For example, another patient with ($isAne: FALSE, isAVM: TRUE, isOcc: FALSE$) would have different probabilities $p_{2}$ and $1 - p_{2}$, where $p_{1} \neq p_{2}$. It is important to note that these probabilities, $p_{1}$ and $p_{2}$, as well as the Siriraj score probabilities, are dependent on the actions. Patients waiting at home (WAIT) tend to have higher uncertainty in these measures, while those hospitalized (HOSP) exhibit lower uncertainty due to constant medical team monitoring.

The Siriraj score, a scalar value ranging from -5 to +5, summarizes the patient's physical and physiological conditions. Scores below -1 typically indicate an ischemic stroke, above 1 suggest a hemorrhagic stroke, and scores in between are considered inconclusive. Our model samples a Siriraj score based on the conditional probability related to the underlying stroke condition. For instance, if $IsAne$ or $IsAVM$ is $TRUE$, the score is more likely to be above 1. Conversely, if $IsOcc$ is $TRUE$, the score tends to be below -1. In the absence of any stroke conditions, the score usually centers around 0.

Both the CT scan summary and the Siriraj score are assumed to be collected for every patient unless a DSA is ordered. In cases where DSA is performed, instead of observing the CT scan and Siriraj score, we obtain binary predictions ($PredAne, PredAVM, PredOcc$) for the underlying state ($IsAne, IsAVM, IsOcc$) with 98\% accuracy. DSA provides reliable estimates but is a costly and invasive procedure, thus reserved for necessary cases, i.e. associated with high cost in the reward function.

\subsection{Belief Model}

The belief model $b_{t}$ represents the probability distribution over the state space $\mathcal S$ at time $t$, given the history of actions and observations. Specifically, for our stroke-related conditions, $b_{t}(s_{t})$ encapsulates the estimated probabilities of state $s_{t}$  based on the noisy observations from CT scans and Siriraj scores. This model is pivotal in scenarios where the actual state st is partially observable, allowing us to infer the most probable state of the patient's condition for decision making.

At each decision epoch, the belief model informs the selection of actions $a_{t} \in \mathcal A$. The action chosen at time $t$, 
$a_{t} = \pi(b_{t})$, is based on the current belief $b_{t}$, where  represents the policy $\pi$ mapping beliefs to actions. This process dynamically adapts to the evolving state of the patient, as each new observation $o_{t}$ and action $a_{t}$ refine the belief model, leading to more informed and effective decision-making over time. Key to this is ways we update the belief in each iteration, especially using particle filters.

\begin{itemize}
\item \textbf{Belief Updating Process:} The belief updating process is a key mechanism in our POMDP model. After executing an action $a_{t}$ and observing an outcome $o_{t}$, the belief state is updated to  $b_{t+1}$ using the Bayes' rule. This update is mathematically represented as \begin{align}
b_{t+1}(s') = \eta ~Z(o_{t} ~| ~s', a_{t}) \sum_{s \in S} T(s' ~| ~s, a_{t})~b_{t}(s),
\label{bayes_rule}
\end{align} where  is a normalizing constant. This formula adjusts the belief distribution to account for the new observation $o_{t}$ and the known effects of the action at, thereby incrementally refining our understanding of the patient's condition despite the inherent noise in the observations. The implementation of this uses particle filters for fast and reliable computation.

\item \textbf{Particle Filters:} Particle filters, in this context, operationalize belief updating through a simulation-based approach. Representing the belief $b_{t}$ as a set of particles $\{s_{t}^{1}, s_{t}^{2}, \dots, s_{t}^{n}\}$, each particle $s_{t}^{i}$ is a hypothesis of the patient's state. Upon receiving a new observation $o_{t}$, the particle filter updates the weights of these particles based on the likelihood $Z(o_{t} | s_{t}^{i}, a_{t})$. The resampling step then focuses on particles with higher weights, effectively concentrating the belief on more probable states. This method is adept at managing the complex dynamics and uncertainties in medical scenarios, offering a nuanced and continuously updated representation of the patient's condition.
\end{itemize}

\section{Experiment Setting}
\label{sec:exp}

Our experiment goal is to compare how a POMDP model described in Section III solved with an online solver performs vs. expert policies and a baseline random policy. We simulate $K = 10,000$ cases with various initial conditions. To this end, 0.785 cases are initialized with stroke-free conditions, 0.05 with only one stroke condition (either aneurysm, AVM, or occlusion), 0.02 with two combinations, and 0.005 with all the conditions. In addition, we assume some random emergence of stroke conditions with probability of $p_{ane}$,  $p_{AVM}$, and $p_{occ}$ for aneurysm, AVM, and occlusion, respectively, during the decision periods. 

The reward function considers the accuracy of treatments for the underlying stroke conditions and impose penalties for mismatch and rewards for matching cases. Table \ref{tab:exp_params} summarizes these parameters, along with other hyperparameters of the experiment. The full detail of the experiment is open-source and available at \href{www.github.com/inteligensi/DSAPOMDPs.jl}{www.github.com/inteligensi/DSAPOMDPs.jl}. 

\begin{table}
  \caption{Problem Parameters}
  \label{tab:exp_params}
  \centering
  \begin{tabular}{>{\raggedright}p{1.7in}ccc}
    \toprule
    Parameter Name & Symbol & Unit & Value \\
    \midrule
    Probability of $IsAne: \text{TRUE}$ with non-treatment action & $p_{ane}$ & - & 0.0005 \\
    Probability of $IsAVM: \text{TRUE}$ with non-treatment action & $p_{AVM}$ & - & 0.0002 \\
    Probability of $IsOcc: \text{TRUE}$ with non-treatment action & $p_{occ}$ & - & 0.0002 \\
    Minimal probability to be considered dominant state prediction (to make treatment action) & $pdom_{thres}$ & - & 0.6 \\
    Minimal prediction confidence of stroke-free state for discharging patients (used by both ExpertHOSP and ExpertDSA policies) & $pdisc_{min}$ & - & 0.9 \\
    Discount factor & $\gamma$ & - & 0.95 \\
    Num. particles for belief updater & $n$ & particles & 100 \\
    Num. replications for performance metrics statistic calculation & $K$ & samples & 10,000 \\
    Penalty when the patient is not treated & - & - & -100,000 \\
    Cost of treatment actions & - & - & -200 \\
    Cost of DSA & - & - & -150 \\
    Cost of hospitalizing patients & - & - & -100 \\
    Reward for correct treatment & - & - & 5,000 \\
    Penalty for wrong treatment & - & - & -5,000 \\
    Reward for needed DSA & - & - & 250 \\
    Penalty for unnecessary DSA & - & - & -750 \\
    Reward for correct HOSP & - & - & 150 \\
    Penalty for unnecessary HOSP & - & - & -400 \\
    Penalty for not hospitalizing stroke patients & - & - & -1,000 \\
    Reward for discharging stroke-free patients & - & - & 5,000 \\
    Penalty for discharging stroke patients & - & - & -50,000 \\
    \bottomrule
  \end{tabular}
\end{table}

We compare four distinct policies: Random, ExpertHOSP, ExpertDSA, and DESPOT, allowing a robust comparison between static, expert-driven policies, and a POMDP strategy. 

\begin{itemize}

\item \textbf{Random policy}: serving as a baseline, involves selecting actions randomly from the action space. This approach, while straightforward, is expected to perform poorly and only establishes a comparative standard against which the more complex policies can be evaluated. 

\item \textbf{ExpertHOSP and ExpertDSA}: grounded in domain expertise, utilize a belief-based decision-making framework. These policies analyze the current belief state to choose the most suitable action. In scenarios where no specific condition is dominant (the probabilities for all states are $\le p_{dom\_thres}$), both policies default to their respective diagnostic actions (HOSP for ExpertHOSP and DSA for ExpertDSA). If the belief in a stroke-free condition is sufficiently high ($> p_{disc\_min}$), the patient is discharged. However, if the probability of a condition surpasses a certain threshold, the corresponding treatment action for the predicted stroke condition with the highest probability is ordered.

\item \textbf{DESPOT}: an advanced online POMDP solver, offers a dynamic and adaptive decision-making strategy. It generates policies by simulating potential future scenarios, making it particularly effective in the complex and uncertain realm of our problem.

\end{itemize}

Our experiment aims to evaluate these policies in terms of recovery rate, time-to-treatment for patients with stroke conditions, and cumulative discounted reward, as detailed in (\ref{optimal_policy}). We analyze the advantages and trade-offs of each approach. The performance metrics achieved by all four policies are summarized in Table \ref{tab:transition_probability}. Figure \ref{fig:hist_rdisc} presents normalized histograms of the overall cumulative discounted reward, while Figure \ref{fig:hist_time2recover} displays histograms of the time-to-treatment for patients with initial stroke conditions under each policy. Figure \ref{fig:hist_rdisc_cases} provides a breakdown of the cumulative discounted reward, differentiating between patients requiring treatment and those who do not. Lastly, we illustrate the example actions taken by all policies for both mild (single stroke type, in this case $IsAne:TRUE$) in Figure \ref{fig:sample_action_one} and severe (multiple stroke types, in this case $IsAVM:TRUE$ and $IsOcc:TRUE$) in Figure \ref{fig:sample_action_two}.

\section{Discussion}
\label{sec:discussion}

In this section, we analyze the performance of all benchmarked policies and examine the sequence of actions for mild and severe cases to demonstrate each policy's strategy.

\subsection{Performance Benchmarks}

Table \ref{tab:performace_metrics} clearly shows that the Random policy is the least effective among all benchmarked policies. Notably, our proposed DESPOT policy performs comparably with both expert policies (ExpertDSA and ExpertHOSP) in terms of overall objective (mean cumulative discounted reward). It also achieves a competitive recovery rate with ExpertDSA, the top performer, and a slightly better time-to-treatment performance than ExpertHOSP. However, due to the large standard deviations in both discounted reward and time-to-treatment metrics across all policies, we cannot claim statistical significance in these performance comparisons. The substantial standard errors underscore the variability in how these algorithms perform across different stroke conditions, including stroke-free conditions, which constitute approximately 78.5\% of the initial patient states.

\begin{table*}
  \caption{Performance Summary}
  \label{tab:performace_metrics}
  \centering
  \begin{tabular}{lcccc}
    \toprule
    Metric & Random & ExpertDSA & ExpertHOSP & DESPOT \\
    \midrule
    Recovery rate & 0.440 & 0.926 & 0.848 & 0.921 \\
    Disc. reward (mean) & -14,053.58 & 3,742.15 & 3,033.46 & 3,105.90 \\
    Disc. reward (std) & 18,657.63 & 9,270.62 & 10,571.92 & 9,098.25 \\
    Time-to-treatment (mean) & 15.26 & 4.41 & 7.37 & 6.31 \\
    Time-to-treatment (std) & 10.10 & 5.61 & 7.66 & 5.60 \\
    \bottomrule
  \end{tabular}
\end{table*}

\begin{figure}
    \centering
    \includegraphics[width=\linewidth]{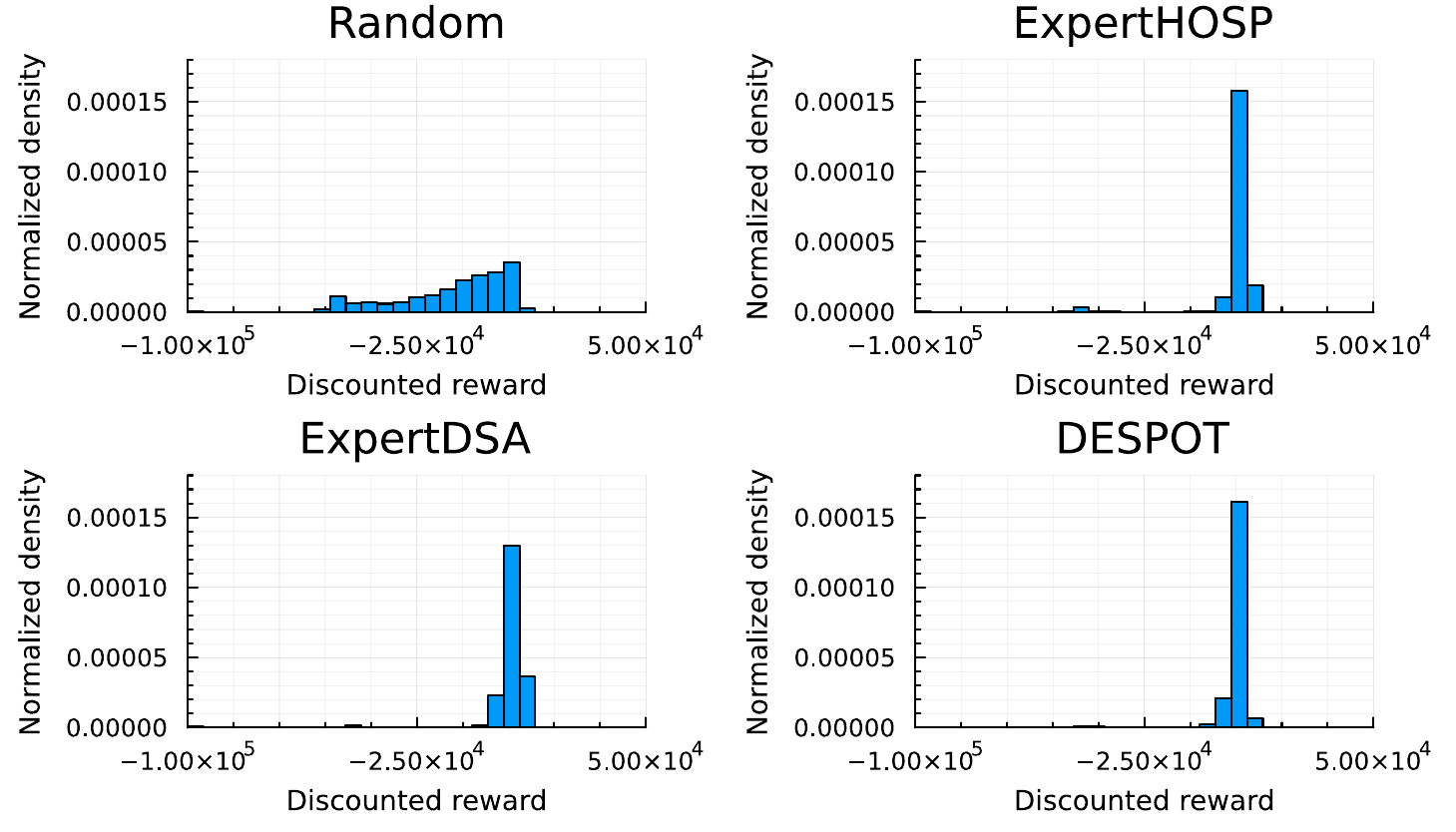}
    \caption{The normalized histograms of discounted reward}
    \label{fig:hist_rdisc}
\end{figure}

\begin{figure}
    \centering
    \includegraphics[width=\linewidth]{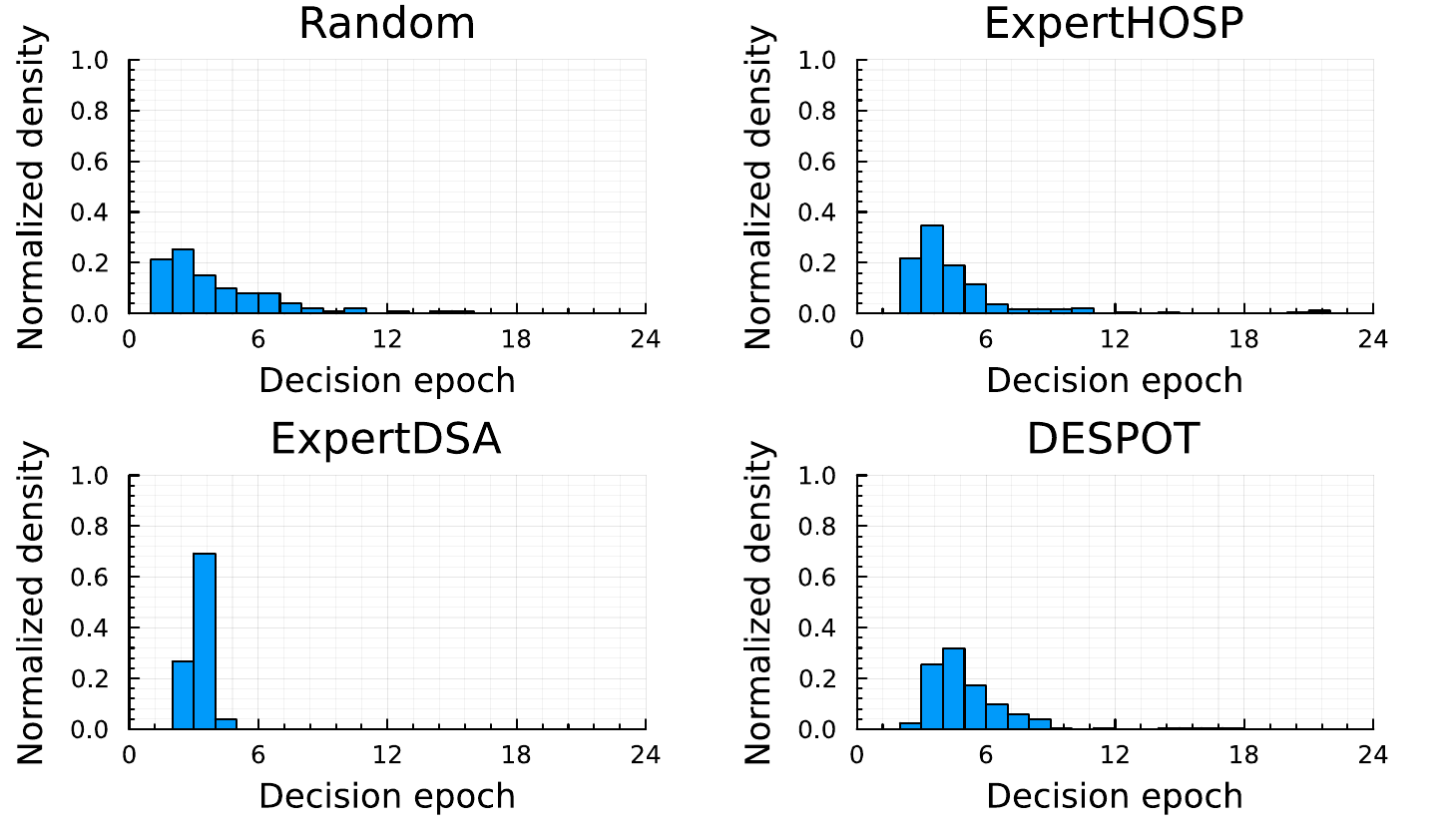}
    \caption{The normalized histograms of time-to-treatment }
    \label{fig:hist_time2recover}
\end{figure}

\begin{figure*}
    \centering
    \includegraphics[width=0.65\linewidth]{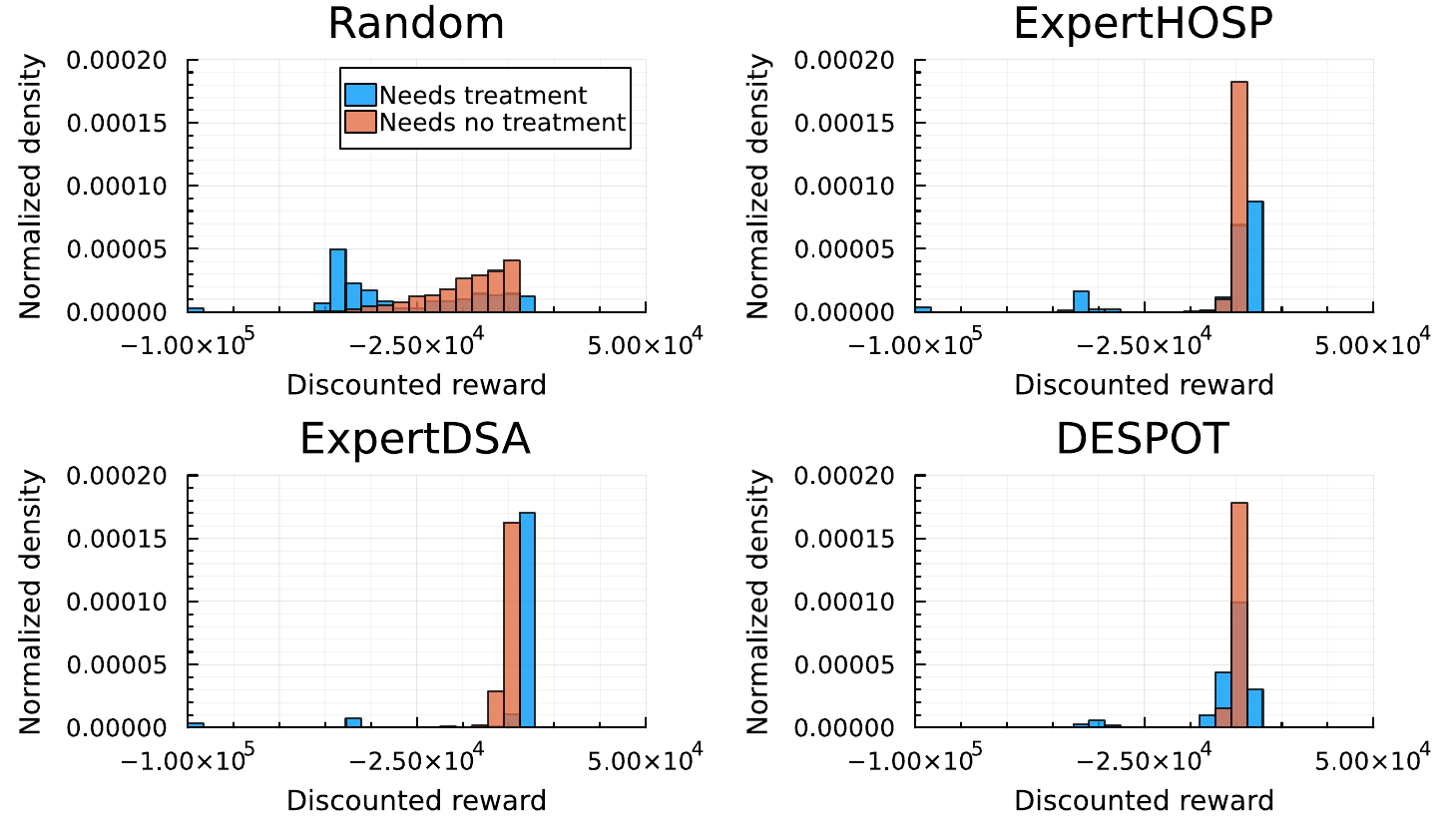}
    \caption{The normalized discounted reward for patient with/without stroke}
    \label{fig:hist_rdisc_cases}
\end{figure*}

Examining the normalized histogram of discounted rewards (Fig. \ref{fig:hist_rdisc}), we observe that the densities of cumulative rewards for ExpertHOSP, ExpertDSA, and DESPOT follow a similar trend, with most densities around ±5,000. However, all policies exhibit some densities at extreme negative values (-20,000 and -50,000), indicating significant penalties from false negative predictions (discharging stroke patients). The Random policy, as expected, displays a distinctly subpar performance pattern.

In terms of time-to-treatment (Fig. \ref{fig:hist_time2recover}), DESPOT and ExpertHOSP show very similar densities, indicating comparable performance. ExpertDSA stands out with its density concentrated on shorter time-to-treatment values, reflecting the decisive nature of DSA in identifying stroke conditions and recommending treatment procedures. Random policy appear to have heavier tails, highlighting the higher likelihood for patients to have time-to-treatment values.

Finally, we further analyze the discounted reward metric by breaking down the density of discounted rewards for both stroke (treatment needed) and stroke-free (no treatment needed) patients (Fig. \ref{fig:hist_rdisc_cases}). This breakdown reveals that while ExpertDSA performs better (higher reward densities) for stroke cases (blue bars), ExpertHOSP slightly outperforms in stroke-free cases (orange bars). DESPOT, relying solely on internal simulations without the inherent expert knowledge of the two expert policies, appears to balance performance between stroke and stroke-free patients.

\subsection{Sampled Actions for Mild and Severe Cases}

\begin{figure}
    \centering
    \includegraphics[width=1\linewidth]{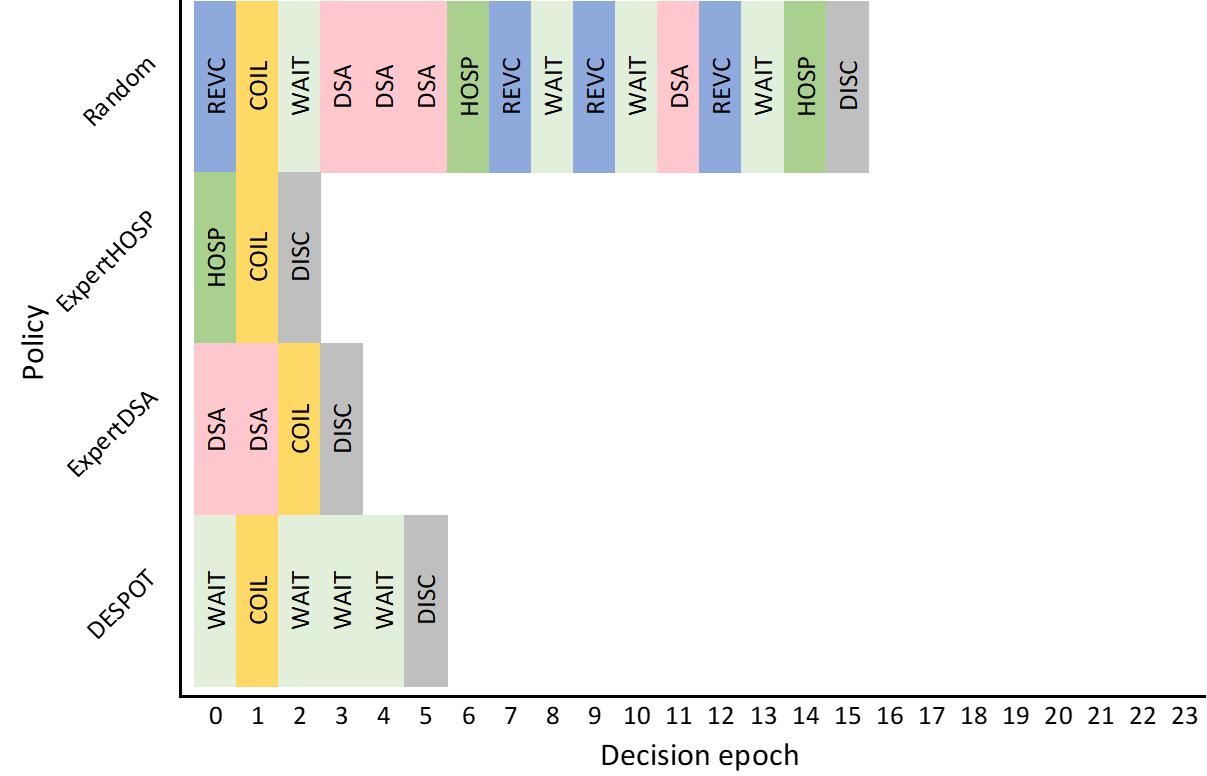}
    \caption{Sampled actions from all policies for a mild case ($IsAne: TRUE, IsAVM: FALSE, IsOcc: FALSE$)}
    \label{fig:sample_action_one}
\end{figure}

We also examined the actions taken by all policies for two different stroke conditions. In a mild case (Fig. \ref{fig:sample_action_one}), where patients exhibit only one stroke type, the Random policy selects arbitrary action sequences, many of which are illogical (e.g., choosing DSA after a REVC procedure). In contrast, ExpertDSA and ExpertHOSP consistently follow their primary strategies, initiating DSA and HOSP actions, respectively, upon patient arrival. They perform treatment (COIL) when the underlying stroke type ($IsAne:TRUE$) is predictable, followed by discharging the patient post-procedure. DESPOT adopts a more cautious and cost-effective approach. It initially recommends a WAIT action to gather CT scan and Siriraj score observations before proceeding with treatment. Interestingly, DESPOT often advises observing the patient with three consecutive WAIT actions to ensure no untreated conditions remain before selecting the DISC action. This approach explains why DESPOT tends to have a longer time-to-treatment compared to ExpertDSA.

\begin{figure}
    \centering
    \includegraphics[width=1\linewidth]{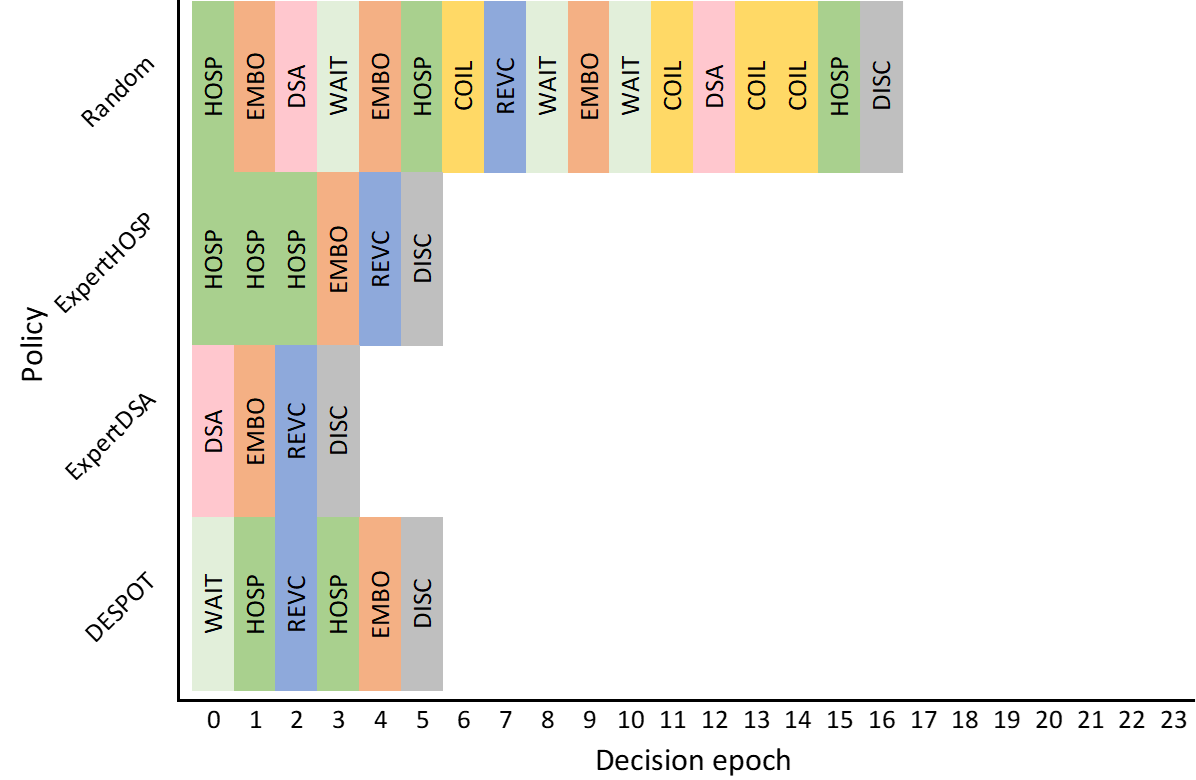}
    \caption{Sampled actions from all policies for a mild case ($IsAne: FALSE, IsAVM: TRUE, IsOcc: TRUE$)}
    \label{fig:sample_action_two}
\end{figure}

Fig. \ref{fig:sample_action_two} illustrates actions for a more severe case (a patient exhibiting both aneurysm and occlusion, occurring with a probability of only 0.02 in our model). In this scenario, DESPOT again adopts a balanced approach, gradually increasing observation accuracy, but only up to the HOSP level (without performing DSA), to build its belief about the underlying stroke conditions. It then performs treatment (REVC) while continuing to gather observations (opting for another HOSP action), before proceeding with another treatment (EMBO) and ultimately discharging the patient. ExpertHOSP and ExpertDSA behave as expected, thoroughly collecting data before consecutively performing treatments. This balanced approach highlights the strength and potential of the POMDP modeling approach in effectively integrating diagnostic and treatment goals, assisting medical professionals in providing safe and cost-effective care amidst uncertainties.

\section{Conclusion}
\label{sec:conclusion}

Our study demonstrates the significant potential of the POMDP modeling approach in enhancing decision-making processes in medical settings, particularly in the context of stroke management. By integrating diagnostic and treatment strategies, the POMDP framework, exemplified by the DESPOT solver, offers a balanced and effective method for navigating the complexities and uncertainties inherent in medical care. This approach not only optimizes patient outcomes by providing tailored treatment plans but also ensures cost-effectiveness and resource optimization. However, our current formulation is not without limitations. One key constraint is the reliance on the accuracy and comprehensiveness of the input data, particularly in representing the diverse range of stroke conditions and patient responses. Looking ahead, future research should focus on incorporating more diverse and extensive datasets, including patient histories and broader stroke-related variables, to refine the model's accuracy and applicability. By addressing these areas, the POMDP framework can be further developed to become an indispensable tool in medical decision-making, ultimately improving patient care and outcomes in various healthcare settings.

\begin{acks}
The authors express their gratitude to the anonymous reviewer for the valuable feedback that has enhanced the quality of this work. This research was partially supported by the Ikatan Ilmuwan Indonesia Internasional (I-4) Research Grant and Faculty of Medicine UIN Syarif Hidayatullah Jakarta, for which we are also deeply thankful. 
\end{acks}

\bibliographystyle{ACM-Reference-Format}
\bibliography{bib}

\end{document}